\documentclass[letterpaper]{article} 
\usepackage{aaai24}  
\usepackage{times}  
\usepackage{helvet}  
\usepackage{courier}  
\usepackage[hyphens]{url}  
\usepackage{graphicx} 
\usepackage{natbib}  
\usepackage{caption} 
\usepackage{amsmath,bm,bbding,color, diagbox}

\usepackage{multicol}
\usepackage{multirow}
\usepackage{arydshln}
\usepackage{mathtools}
\usepackage{algorithm}
\usepackage{algorithmic}
\newcommand{\eqdef}{\stackrel{\text{def}}{=}}
\usepackage{newfloat}
\usepackage{listings}
\DeclareCaptionStyle{ruled}{labelfont=normalfont,labelsep=colon,strut=off} 
\floatstyle{ruled}
\newfloat{listing}{tb}{lst}{}
\floatname{listing}{Listing}
\title{BiPFT: Binary Pre-trained Foundation Transformer with Low-Rank Estimation of Binarization Residual Polynomials}
\author{
    Xingrun Xing\textsuperscript{\rm 1,2,3},
    Li Du\textsuperscript{\rm 3},
    Xinyuan Wang\textsuperscript{\rm 4},
    Xianlin Zeng\textsuperscript{\rm 4},
    Yequan Wang\textsuperscript{\rm 3},
    Zheng Zhang\textsuperscript{\rm 3}\thanks{Corresponding Author},\\
    Jiajun Zhang\textsuperscript{\rm 1}\footnotemark[1]
}
\affiliations{
    \textsuperscript{\rm 1}Institute of Automation, Chinese Academy of Sciences\\
    \textsuperscript{\rm 2}School of Artificial Intelligence, University of Chinese Academy of Sciences\\
    \textsuperscript{\rm 3}Beijing Academy of Artificial Intelligence\\
    \textsuperscript{\rm 4}Beihang University\\

    xingxingrun2023@ia.ac.cn, duli@baai.ac.cn, buaa42wxy@gmail.com, zengxianlin@buaa.edu.cn, tshwangyequan@gmail.com, zhangz.goal@gmail.com, jjzhang@nlpr.ia.ac.cn
}

\usepackage{bibentry}

\begin{document}

\maketitle

\begin{abstract}
Pretrained foundation models offer substantial benefits for a wide range of downstream tasks, which can be one of the most potential techniques to access artificial general intelligence. However, scaling up foundation transformers for maximal task-agnostic knowledge has brought about computational challenges, especially on resource-limited devices such as mobiles. This work proposes the first Binary Pretrained Foundation Transformer (BiPFT) for natural language understanding (NLU) tasks, which remarkably saves 56$\times$ operations and 28$\times$ memory. In contrast to previous task-specific binary transformers, BiPFT exhibits a substantial enhancement in the learning capabilities of binary neural networks (BNNs), promoting BNNs into the era of pre-training. 
Benefiting from extensive pretraining data, we further propose a data-driven binarization method.
Speciﬁcally, we ﬁrst analyze the binarization error in self-attention operations and derive the polynomials of binarization error.
To simulate full-precision self-attention, we define binarization error as binarization residual polynomials, and then introduce low-rank estimators to model these polynomials.
Extensive experiments validate the effectiveness of BiPFTs, surpassing task-specific baseline by 15.4\% average performance on the GLUE benchmark.
BiPFT also demonstrates improved robustness to hyperparameter changes, improved optimization efficiency, and reduced reliance on downstream distillation, which consequently generalize on various NLU tasks and simplify the downstream pipeline of BNNs. Our code and pretrained models are publicly available at https://github.com/Xingrun-Xing/BiPFT.
\end{abstract}

\section{Introduction}

\begin{figure}[t]
\centering
\includegraphics[width=1\columnwidth]{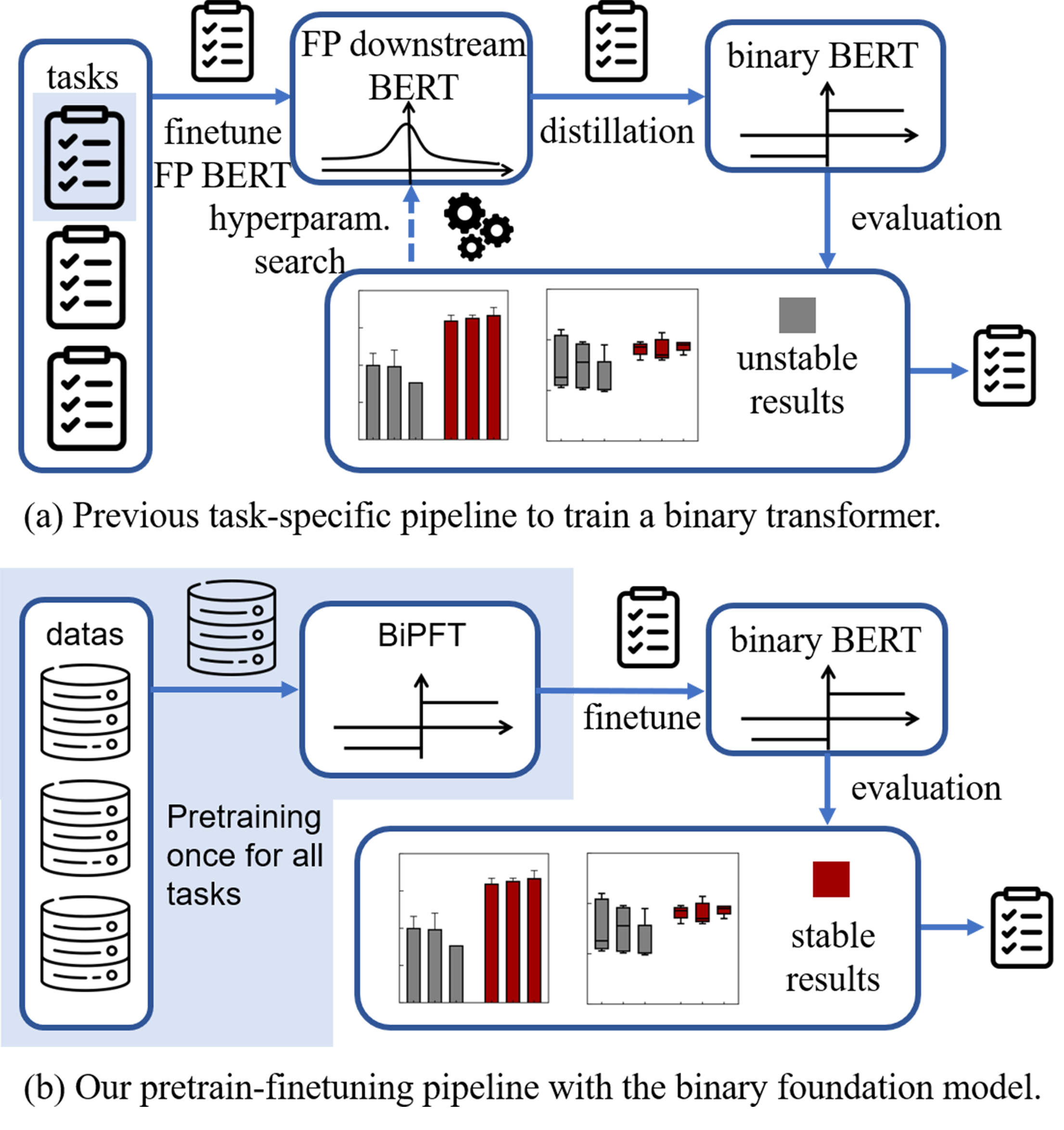}
\caption{Comparison of training pipelines for binary transformers. FP indicates full-precision. For downstream tasks, finetuning BiPFT replaces previous task-specific pipelines.}
\label{figure1}
\end{figure}

In recent years, pre-trained foundation models (PFM) \cite{open111} have demonstrated impressive emergent intelligence phenomena in various fields such as natural language processing \cite{touvron2023llama} and computer vision \cite{kirillov2023segment, wang2023seggpt}. As the model size and pre-training data increase, task-agnostic knowledge from pretraining effectively generalizes to downstream tasks with small datasets or open scenes. In natural language understanding (NLU) tasks, BERT \cite{devlin2018bert, he2023debertav}, which uses the transformer encoder architecture and the bi-directional masked prediction training, is widely applied. However, the self-attention \cite{vaswani2017attention, carlini2023quantifying} and MLP layers in BERT involve substantial floating-point operations and memory consumption. How to get a compact pretrained foundation model in computationally limited settings, such as inference at mobile devices, has become a problem of significant value.

This work aims to propose the first 1-bit pretrained foundation model for NLU tasks with a BERT like architecture. Recently, compression methods for BERT include model pruning \cite{gordon2020compressing, zhao2023holistic}, distillation \cite{sun2020mobilebert, ding2023skdbert}, and quantization \cite{kim2021bert, castano2023equivalence}. Model quantization achieves a high degree of model compression without changing the model architecture or the number of parameters. Notably, 1-bit model quantization is an extreme case of low-bit quantization. Unlike other low-bit models, binary neural networks (BNNs) \cite{courbariaux2016binarized, xu2023resilient} directly utilize the underlying XNOR and popcount operations instead of numerical ones, thereby achieving \emph{super-linear benefits of bit-width}. Compared to full-precision (FP) model inference, binary models save up to 64$\times$ operations, 32$\times$ memory, and between 100 to 1000$\times$ energy consumption \cite{courbariaux2016binarized}, which are necessary for modern large pretrained foundation models.

Current binary transformers perform binarization on specific tasks. Due to their extremely low bit-width, these binary transformers face significant optimization challenges. To address this issue, prior binary BERTs rely on optimization techniques such as distillation from the full-precision (FP) teacher and hyperparameter tuning. As illustrated in Fig. \ref{figure1} (a), the typical pipeline is complex: it begins by training a FP teacher for the given task, followed by initializing and distilling the binary BERT using this FP teacher. Because of unstable optimization, a hyperparameter search is usually necessary.
We want to ask the question \emph{whether a 1-bit BERT, with initialization and distillation from the downstream FP model, is able to achieve similar performance even without pretraining?}
We build a strong binary transformer baseline and conduct extensive experiments in different training settings (including the distillation, learning rate, batch size, etc.) and want to find the keypoints to influence performance.
These non-trivial experiments indicate the weakness of task-specific binary transformers:

\noindent\textbf{Unstability to hyperparameters.} Our experiments show that task-specific binary BERTs have a large performance variance to different batch sizes and learning rates. The performance heavily relies on hyperparameters tuning and often requires a small batch size and long training time.

\noindent\textbf{Weakness of learning capabilities.} Existing task-specific binary BERTs \cite{qin2022bibert, liu2022bit} also heavily rely on the distillation from FP teacher. When we replace the distillation loss with direct training loss, the average performance drops by 13.9\% on the GLUE benchmark.

\noindent This phenomenon suggests the necessity of directly training 1-bit foundational models, rather than initializing a binary model with its 32-bit task-specific counterpart.

We propose the first Binary Pretrained Foundation Transformer, termed BiPFT, promoting BNNs into the era of pre-training. We start with building a general baseline architecture for binary transformers. Based on this architecture, we then pretrain a binary foundation model named BiPFT-A and evaluate the impact of pretraining for BNNs.
During pretraining, we followed the standard masked language model (MLM) and next sentence prediction (NSP) tasks used in FP BERTs. In addition, a task-agnostic distillation is also attached to speed up pretraining. In contrast to task-specific distillation, task-agnostic distillation doesn't complicate the downstream pipeline. After pretraining, the learning capabilities of binary transformers are improved significantly, which enables binary transformers directly finetuned without distillation and hyperparameter tuning. As shown in Fig. \ref{figure1} (b), given a new task, binary pretrained foundation transformers only require straightforward finetuning, eliminating the complexity of previous downstream pipelines. 
Experimental results show that under fair comparison, BiPFT-A improves 13.9\% average performance on the GLUE benchmark compared with the baseline model without binary pretraining. Even when compared to the baseline that employs additional hyperparameter tuning and distillation, BiPFT-A still surpasses it by 1.1\% with simple finetuning.

With the pretraining phase, we rethink how to effectively binarize self-attention. 
Previous works mainly focus on empirically designing more accurate binary operations. For example, BiBERT \cite{qin2022bibert} designs a Bi-Attention operation to simulate FP self-attention; BiT \cite{liu2022bit} designs the \{0, 1\} binarization level and elastic binarization functions to better simulate FP activations. 
In contrast to performing binarization in downstream tasks with very limited data previously, binary pretrained foundation models perform binarization in the pretraining phase, making it possible to use data-driven and data-hungry binarization methods.
Specifically, we analyze the binarization error in self-attention operations and derive the polynomials of binarization error. To simulate full-precision self-attention, we indicate binarization error as binarization residual polynomials and then introduce low-rank estimators to model binarization residual polynomials.
Low-rank estimators are fully trained in pretraining, while estimators generalize to data-limited tasks effectively in downstream.
We add the aforementioned residual polynomial estimators to BiPFT-A and name the new model as BiPFT-B. Experimental results indicate that BiPFT-B enhances performance on GLUE by an additional 1.6\% compared to BiPFT-A. 

The contributions of this paper are as follows:
\begin{itemize}
\item We propose the first binary pretrained foundation model and successfully train BNNs throughout the pretraining and finetuning phases.
\item We propose a data-driven binarization method for self-attention by estimating binarization residual polynomials, further improving binary foundation models.
\item We release binary foundation transformers for NLU tasks. Finetuning on this foundation model for downstream tasks significantly simplifies the training process of BNNs, yielding more robust and accurate results.
\end{itemize}

\section{Related Work}

Most studies of binary neural networks \cite{he2023deep, kunes2023gradient, bipose} focus on convolutional neural networks in the computer vision field. BNNs are first proposed by directly binarizing both activations and weights to the bit-width of 1 and estimating gradient using straight-through estimators (STE) \cite{courbariaux2016binarized}. However, vanilla BNNs encounter performance drop in large-scale datasets. Many works improve BNN performance from different perspectives, including model architecture, binary parameter optimization and binarization strategy \cite{martinez2020training}. For example, BiRealNet \cite{liu2018bi} and CP-NAS \cite{li2020cp} revise more efficient binary network architectures. XNOR-Net and Siman \cite{lin2022siman} focus on optimizing binarization error. ReActNet \cite{liu2020reactnet} revise binarization and activation functions to improve model capacity. More recently, BCDNet \cite{xing2022towards} introduce MLP \cite{chen2023cyclemlp} architecture to BNNs and achieve high performance.

In the natural language processing field, BinaryBERT \cite{bai-etal-2021-binarybert}, BiBERT \cite{qin2022bibert} and BiT \cite{liu2022bit} binarize full-precision BERT model in specific tasks. TBT \cite{liu2023binary} and DQ-BART \cite{li2022dq} distill binary and low-bit generation models in specific tasks.
However, previous binary transformers heavily rely on task-specific distillation. 
There is no foundation model directly pretrained with binary parameters and activations. 

Compared with post-training quantization (PTQ), BNNs adopt quantization-aware training (QAT). Although some PTQ methods are well-known for language models such as OPTQ \cite{frantar2022optq} and SmoothQuant \cite{xiao2023smoothquant}, they cannot achieve 1-bit width.

\section{Methodology}
\subsection{Build Binary Baseline Architecture}

We define a baseline model as the benchmark of binary transformers and introduce pretraining of the baseline model in the next section. Existing task-specific binary transformers often use different binarization, training and evaluation methods, making it challenging to compare the general performance. 
To build a general baseline, we follow the binarization design of BiTs as much as possible, while replacing their specific training and evaluation settings with common ones.
The differences between our baseline and BiTs are shown in the Appendix A of our extended version \cite{xing2023bipft} in detail. We briefly introduce the basic binary operations in the baseline as follows:

\noindent\textbf{Binary linear.} Binary linear layers compose the most basic operations in a binary transformer, which indicates binarizing both weights and activations to the bit-width of 1. In forward propagation, FP weights $\mathbf{W}$ and activations $\mathbf{A}$ are initially binarized to $\mathbf{W_B}$ and $\mathbf{A_B}$ using the binarization function $\mathbf{Q_B}$. Consequently, linear layers can be carried out as a matrix multiplication with XNOR and popcounting ($\otimes$):
\begin{equation}  \label{eq_linear}
\operatorname{Linear}(\mathbf{A})\approx{\mathbf{Q_B}(\mathbf{W})}\mathbf{Q_B}(\mathbf{A})=\alpha (\mathbf{W_B}\otimes \mathbf{A_B}).
\end{equation}
The simplest $\mathbf{Q_B}$ is the symbolic function, $\operatorname{sign}$:
\begin{equation}
\mathbf{Q_B}(x)=\operatorname{{sign}}(x)= \begin{cases}-1, & \text { if } x<0 \\ +1, & \text { if } x \ge 0\end{cases}, 
\end{equation}
In backward propagation, gradient can't be directly calculated through the $\operatorname{sign}$. Straight-through estimators (STE) are used to estimate gradient:
\begin{equation}
\frac{\partial \operatorname{sign}(x)}{\partial x}\approx \begin{cases}1 & \text { if }|x| \leq 1 \\ 0 & \text { otherwise }\end{cases}.
\end{equation}
Many works make efforts to find more effective binarization functions. BiTs binarize weights by $\mathbf{Q}_{\mathbf{B},w}$, and binarize activations by $\mathbf{Q}_{\mathbf{B},a}$ respectively:
\begin{equation}  \label{eq_1}
\mathbf{Q}_{\mathbf{B},w}^{(-1,+1)}(\mathbf{W})=\frac{\left\|\mathbf{W}\right\|_{l 1}}{n_{\mathbf{W}}}{\operatorname{sign}}\left(\mathbf{W}-\overline{\mathbf{W}}\right),
\end{equation}
\begin{equation}  \label{eq_2}
\mathbf{Q}_{\mathbf{B},a}^{(-1,+1)}(\mathbf{A})=\alpha{\operatorname{sign}}\left(\mathbf{A}-\beta\right),
\end{equation}
where $\alpha, \beta$ are trainable parameters. Omitting scaling factors, both weights and activations have the same binarization level $\{-1, +1\}$. Different from BiTs, we remove the $\{0, 1\}$ binarization level in linear layers because different binarization levels need special transformation to avoid the ternary value problem. All linears are binarized as Eq.\ref{eq_1}, \ref{eq_2} in our general baseline.

\noindent\textbf{Binary self-attention.}
FP self-attention is defined as cascaded matrix productions between the query, key and value:
\begin{equation}  \label{eq_3}
\operatorname{Attention}(\mathbf{Q}, \mathbf{K}, \mathbf{V})=\operatorname{softmax}\left(\frac{\mathbf{Q} \mathbf{K}^T}{\sqrt{d_k}}\right) \mathbf{V}.
\end{equation}
Binary self-attention also consists of two steps: calculating self-attention map, $\operatorname{\mathbf{Att}}$, and reweight value with binarized attention map respectively:
\begin{equation}  \label{eq_4}
\operatorname{\mathbf{Att}} \approx \operatorname{softmax}\left(\frac{\mathbf{Q}_{\mathbf{B},a}^{(-1,+1)}(\mathbf{Q}) \mathbf{Q}_{\mathbf{B},a}^{(-1,+1)}(\mathbf{K^T})}{\sqrt{d_k}}\right),
\end{equation}
\begin{equation}  \label{eq_5}
\operatorname{Attention}(\mathbf{Q}, \mathbf{K}, \mathbf{V})\approx \mathbf{Q}_{\mathbf{B},a}^{(0,+1)}(\mathbf{Att}) \mathbf{Q}_{\mathbf{B},a}^{(-1,+1)}(\mathbf{V}),
\end{equation}
where the binarization function for the attention map is defined as follows in BiTs: 
\begin{equation}
\mathbf{Q}_{\mathbf{B},a}^{(0,+1)}(\mathbf{A})=\alpha\left\lfloor\operatorname{Clip}\left(\frac{\mathbf{A}-\beta}{\alpha}, 0,1\right)\right\rceil,
\end{equation}
\begin{equation}  \label{eq_8}
\left\lfloor\operatorname{Clip}\left(x, 0,1\right)\right\rceil= \begin{cases}0, & \text { if } x<0.5 \\ 1, & \text { if } x \ge 0.5\end{cases}. \nonumber 
\end{equation}
After binarization, values in the attention map become $\{0,1\}$ and formulate hard attention (omitting scaling factors). However, in Eq. \ref{eq_5}, matrix production between $\mathbf{Att_B} \in \{0,+1\}^n$ and $\mathbf{V_B} \in \{-1,+1\}^n$ can't directly be implemented by XNOR and popcount at inference, which needs ternary operations in domain $\{-1,0,+1\}$. It consumes double binary operations to transform ternary to binary operations:
\begin{equation}
\mathbf{Att_{(0,1)}}\mathbf{V_B} = (\mathbf{Att_B}\otimes \mathbf{V_B}+\mathbf{1}\otimes \mathbf{V_B})>>1,
\end{equation}
where $\mathbf{Att_B},\mathbf{V_B} \in \{-1,+1\}^n$, $>>$ is bitshift, and $\mathbf{Att_B}$ is constructed by directly replacing 0 as -1 in $\mathbf{Att_{(0,1)}}$.

\subsection{Pretrain Binary Transformers}
In this section, we propose pretrained foundation transformers based on the baseline architecture, termed BiPFT-A. We use simple but efficient pretraining tasks following the vanilla BERT and task-agnostic distillation \cite{wang2020minilm}. The pretraining tasks in BERT include masked language model and next sentence prediction. Additionally, inspired by the phenomenon that task-agnostic distillation improves pretraining efficiency in small models, we add distillation loss for both token and sentence-level features. In summary, the pretraining objectives of BiPFTs include:

\noindent\textbf{Masked Language Model} ($\ell_{\mathrm{MLM}}$): MLM objective is defined as minimizing the cross-entropy loss between the real and the prediction of masked tokens. Following BERTs, we randomly select 15\% of the input tokens. Among these chosen tokens, 80\% are swapped with [MASK], 10\% are maintained as they are, and the remaining 10\% tokens are substituted with a token randomly picked from the vocabulary. 

\noindent\textbf{Next Sentence Prediction} ($\ell_{\mathrm{NSP}}$): NSP is defined as a binary classification task, where the objective is to predict if two segments appear consecutively in the source text. Following BERTs,  we construct positive samples by selecting sequential sentences from the text corpus and negative samples are by pairing sentences from separate documents. The probability of positive and negative samples is equal.

\noindent\textbf{Task-agnostic Distillation} ($\ell_{\mathrm{logit}}$, $\ell_{\mathrm{rep}}$): previous works \cite{hinton2015distilling} has shown minimizing KL divergency between model logits of the student and teacher achieves better performance than direct training. Following task-agnostic distillation \cite{sun2020mobilebert, wang2020minilm}, we distill logits in the last layer during pretraining. To improve convergency, we additionally apply L2 loss to distill hidden states layer by layer.

\noindent We use the aforementioned objectives to jointly train binary transformers in extensive pretraining data:
\begin{equation} 
\ell_{\text {total }}=\ell_{\mathrm{MLM}}+\ell_{\mathrm {NSP }}+ \frac{1}{n}\sum_{i=1}^{n} \ell_{\text {rep }}^i+\ell_{\text {logit }}.
\end{equation}

After pre-training, task-agnostic knowledge significantly enhances the learning ability of the baseline models in various downstream tasks. As shown in Fig. \ref{figure1} (b), once pretraining finished, we finetune the binary foundation model in various downstream tasks the same as full-precision cases, which bridges the training gap between full-precision and binary foundation models. 

\begin{table}[t]
\centering
\begin{tabular}{l|c|cc|c}
\hline
method & BP & dist. & HS & binarization \\
\hline
BiT  & \XSolidBrush & \Checkmark & \Checkmark & direct binarization \\
\hline
baseline$^*$ & \XSolidBrush & \Checkmark & \Checkmark & direct binarization\\
baseline &\XSolidBrush & \XSolidBrush & \XSolidBrush & direct binarization\\
BiPFT-A & \Checkmark & \XSolidBrush &\XSolidBrush & direct binarization\\ 
BiPFT-B & \Checkmark & \XSolidBrush &\XSolidBrush & error estimatation\\
\hline
\end{tabular}

\caption{Summarization of proposed models, where BP indicates binary pretraining; dist. indicates task-specific distillation; HS indicates hyperparameter search in specific tasks.}
\label{methods}
\end{table}

\subsection{Estimate Binarization Polynomials}

With the help of the pretraining phase, we explore how to better simulate self-attention with binary representations. 
To make full use of pretraining data, we investigate data-driven binarization methods.
In this section, we first analyze binarization errors in self-attention and then propose binarization error estimators to achieve accurate binary self-attention. We add binarization error estimators to baseline architecture and pretrain this binary transformer as BiPFT-B. A summarization of proposed models is shown in Table \ref{methods}.

Self-attention involves cascaded multiplications, making it challenging for previous empirical binarization designs:

\noindent\textbf{Dynamic binary value.} Previous BNNs focused on a more accurate simulation of matrix multiplication between real-valued weights and activations, where activations are dynamic values changing with input, and weights are fixed parameters. However, in self-attention, both items of matrix multiplication are dynamic values changing with inputs.

\noindent\textbf{Cascaded multiplications.} Self-attention has cascaded matrix multiplications. The error accumulation caused by direct binarization affects the accuracy of binary features. For instance, binarization errors from the matrix multiplication of keys and queries undoubtedly impact the following reweight between attention scores and values.

\begin{table*}[t]
\begin{center}
\renewcommand\arraystretch{0.7}
\centering
\setlength{\tabcolsep}{1.2mm}
\fontsize{9.3pt}{\baselineskip}\selectfont
\begin{tabular}{lcccccccccccc}
\hline
\textbf{Quant} & \begin{tabular}[c]{@{}c@{}}\textbf{E-W-A}\end{tabular} & \begin{tabular}[c]{@{}c@{}}\textbf{Size}$_\text{ (MB)}$\end{tabular} & \begin{tabular}[c]{@{}c@{}}\textbf{FLOPs}$_\text{ (G)}$\end{tabular}  & \textbf{STS-B} & \textbf{MRPC} & \textbf{RTE} & \textbf{QQP} & \textbf{QNLI} & \textbf{SST-2} & \textbf{COLA} & \begin{tabular}[c]{@{}c@{}}\textbf{MNLI}$_\text{-m/mm}$\end{tabular} & \textbf{Avg.} \\ 
\hline
BERT$_{base}$  & 32-32-32 & 418 & 22.5 & 88.8 & 86.5 & 66.8 & 91.4 & 91.3 & 92.9 & 55.8 & 83.4/83.6  & 82.3 \\
Q-BERT         & 2-8-8    & 43.0 & 6.5 & --   & 68.3 & 52.7 & --   & --   & 84.6 & --   & 76.6/77.0  & --   \\
Q2BERT         & 2-8-8    & 43.0 & 6.5 & 4.4  & 68.4 & 52.7 & 67.0 & 61.3 & 80.6 & 0    & 47.2/47.3 & 47.7 \\
{TernaryBERT}  & 2-2-8    & 28.0 & 6.4 & --   & 87.5 & 68.2 & 90.1 & --   & --   & 50.7 & 83.3/83.3 & --   \\
{BinaryBERT}   & 1-1-8    & 16.5 & 3.1 & 88.6 & 85.5 & 72.2 & 91.2 & 91.5 & 92.6 & 53.4 & 84.2/84.7 & 82.7 \\
{BinaryBERT}        & 1-1-4    & 16.5 & 1.5 & 87.2 & 83.3 & 65.3 & 91.2 & 90.9 & 92.3 & 44.4 & 83.9/84.2 & 79.9 \\
\hline
\multicolumn{3}{l}{\textit{Finetuning with task-specific distillation}} &&&&&&&&&& \\
{BinaryBERT}        & 1-1-1 & 16.5 & 0.4 & 6.1  & 68.3 & 52.7 & 66.2 & 51.5 & 53.2 & 0    & 35.6/35.3 & 41.0 \\
{BiBERT}            & 1-1-1 & 13.4 & 0.4 & 33.6 & 72.5 & 57.4 & 84.8 & 72.6 & 88.7 & 25.4 & 66.1/67.5 & 63.2 \\
{Baseline$^*$}      & 1-1-1 & 14.7 & 0.4 & 53.4 & 76.0 & 56.7 & 85.5 & 84.2 & 85.7 & 21.9 & 74.8/75.4 & 68.1 \\
\hdashline[0.8pt/1pt]
\multicolumn{5}{l}{\textit{Finetuning without task-specific distillation}} &&&&&&&& \\
{Baseline}          & 1-1-1 & 14.7 & 0.4 & 19.7 & 70.3 & 57.0 & 78.0 & 60.2 & 79.7 & 16.5 & 58.8/58.7 & 55.4 \\
BiPFT-A             & 1-1-1 & 14.7 & 0.4 & 79.0 & 74.0 & 60.6 & 82.8 & 80.3 & 85.6 & 19.8 & 70.3/70.8 & 69.2 \\
BiPFT-B             & 1-1-1 & 14.9 & 0.4 & 80.2 & 76.2 & 66.1 & 83.7 & 81.7 & 86.2 & 22.9 & 69.5/70.6 & 70.8 \\
\hline
BinaryBERT          & 1-1-2 & 16.5 & 0.8 & 6.5  & 68.3 & 52.7 & 79.9 & 52.6 & 82.5 & 14.6 & 62.7/63.9 & 53.7 \\
BiT                 & 1-1-2 & 14.7 & 0.8 & 82.2 & 78.4 & 58.1 & 87.1 & 89.3 & 90.8 & 32.1 & 82.1/82.5 & 75.0 \\
BiPFT-B             & 1-1-2 & 14.9 & 0.8 & 87.0 & 84.1 & 66.1 & 89.0 & 86.6 & 88.1 & 36.2 & 77.0/76.9 & 76.8 \\
\hline
\end{tabular}
\caption{Comparison of BERT quantization methods on the GLUE dev set. The E-W-A refers to the bit-width of embeddings, weights and activations. The baseline and baseline$^*$ are described in Table 1, which have almost the same architecture as BiT but evaluated in our common settings.}
\label{glue}
\end{center}
\end{table*}

To find where binarization errors occur in self-attention, we first compare the differences before and after binarization; then define the residual polynomials ignored previously; and finally, we use low-rank estimators to model these residuals.
In order to decompose the binarization error, we define binarization residuals of the query, key, and value items as well as their weights:
\begin{eqnarray}  \label{eq_12}
\mathbf{Q^{*}} \eqdef \mathbf{Q}-\mathbf{Q_B}, \: &\mathbf{K^{*}} \eqdef \mathbf{K}-\mathbf{K_B}&, \: \mathbf{V^{*}} \eqdef \mathbf{V}-\mathbf{V_B},  \nonumber \\
&\mathbf{W}^{*} = \mathbf{W}-\mathbf{W_B}&.
\end{eqnarray}
According to Eq. \ref{eq_3}, we first focus on the attention score between keys and queries. The full-precision $\mathbf{Q}$ and $\mathbf{K}$ can be decomposed into the sum of their binarized parts, $\mathbf{Q_B}$ and $\mathbf{K_B}$, and their binarization residuals, $\mathbf{Q^{*}}$ and $\mathbf{K^{*}}$. As shown in Eq. \ref{eq_11}, in the simplified polynomials, the first term can be represented as directly binarized multiplication between $\mathbf{Q}$ and $\mathbf{K}$, while the other three terms constitute the quantization error:
\begin{eqnarray}  \label{eq_11}
\mathbf{A_{score}}&=&\mathbf{Q}\mathbf{K^T} \\
~&=&(\mathbf{Q_B}+\mathbf{Q^{*}})(\mathbf{K_B^T}+\mathbf{K^{* T}}) \nonumber \\
~&=&\mathbf{Q_B}\mathbf{K_B^T}+\underbrace{\mathbf{Q_B}\mathbf{K^{* T}}+\mathbf{Q^{*}}\mathbf{K_B^T}+\mathbf{Q^{*}}\mathbf{K^{* T}}}_{residual \: polynomials}. \nonumber
\end{eqnarray}
Previous binary operations are mainly designed for linear or convolutional layers in computer vision, with a lack of consideration for the multiplication between activations in self-attention. Directly replacing real matrix multiplication with binarized matrix multiplication after quantization overlooks the residual polynomials in Eq. \ref{eq_11}, leading to binarization errors. Towards accurate binary self-attention, we propose data-driven estimators to model these binarization residual polynomials. We indicate residual polynomials in Eq. \ref{eq_11} as $\mathbf{A_{score}^{*}}$ and model these items by low-rank estimators:
\begin{eqnarray}  \label{eq_18}
\mathbf{A_{score}^{*}}&=&\mathbf{A}\mathbf{W_q}\mathbf{W_k^{* T}}\mathbf{A^T}+\mathbf{A}\mathbf{W_q^{*}}\mathbf{W_k^{T}}\mathbf{A^T} \nonumber \\
~&+&\mathbf{A}\mathbf{W_q^{*}}\mathbf{W_k^{* T}}\mathbf{A^T} \nonumber \\
~&\approx&\mathbf{A}\mathbf{w_q}\mathbf{w_k^{* T}}\mathbf{A^T}+\mathbf{A}\mathbf{w_q^{*}}\mathbf{w_k^{T}}\mathbf{A^T} \nonumber \\
~&+&\mathbf{A}\mathbf{w_q^{*}}\mathbf{w_k^{* T}}\mathbf{A^T},
\end{eqnarray}
where $\mathbf{W_{q,k}^{(*)}} \in \mathbf{R}^{\mathbf{C \times C}}$, $\mathbf{w_{q,k}^{(*)}} \in \mathbf{R}^{\mathbf{C} \times 1}$ and $\mathbf{C}$ donates hidden size of the transformer. In Eq. \ref{eq_18}, $\mathbf{w_{q,k}^{(*)}}$ are trainable parameters and used as approximations of $\mathbf{W_{q,k}^{(*)}}$ respectively. In that case, the original dense matrix multiplications $\mathbf{A}\mathbf{W_{q,k}^{(*)}}$ are approximated as low-rank multiplications $\mathbf{A}\mathbf{w_{q,k}^{(*)}}$. We set the rank number as 1 to save 768$\times$ operations in the base-sized BERT, which will not introduce much additional cost. 
As a result, low-rank multiplications approximate residual polynomials ignored by direct binarization. 
In BiPFT-B, Eq. \ref{eq_4} in the baseline model is replaced by Eq. \ref{eq_24}:
\begin{equation}  \label{eq_24}
\operatorname{\mathbf{Att}} \approx \operatorname{softmax}\left(\frac{\mathbf{Q}_{\mathbf{B}} \mathbf{K_B^T}+\mathbf{A_{score}^{*}}}{\sqrt{d_k}}\right).
\end{equation}

We apply a similar analysis to the reweight multiplication in Eq. \ref{eq_5}. We decompose full-precision value $\mathbf{V}$ into binary $\mathbf{V_B}$ and its resdual $\mathbf{V^{*}}$. Binarization residual polynomial can be represented as $\mathbf{Att}(\mathbf{A}\mathbf{W_v^{*}})$:
\begin{eqnarray}  \label{eq_26}
\operatorname{Attention}(\mathbf{Q},\mathbf{K},\mathbf{V})&=&\mathbf{Att}\mathbf{V} \\
~&=&\mathbf{Att}(\mathbf{V_B}+\mathbf{V^{*}}) \nonumber \\
~&=&\mathbf{Att}\mathbf{V_B}+\mathbf{Att}(\mathbf{A}\mathbf{W_v^{*}}). \nonumber
\end{eqnarray}
In contrast to decomposing attention map, $\mathbf{Att}$, we directly use binarized attention map, $\mathbf{Att_B}$, because binary attention map formulates hard attention that we don't want to break. In BiPFT-B, Eq. \ref{eq_5} in the baseline is replaced by Eq. \ref{eq_27}:
\begin{eqnarray}  \label{eq_27}
\operatorname{Attention}(\mathbf{Q},\mathbf{K},\mathbf{V})\approx\mathbf{Att_B}\mathbf{V_B}+\mathbf{Att_B}(\mathbf{A}\mathbf{w_v^{*}}).
\end{eqnarray}

\section{Experiments}

\subsection{Experiment Settings}
In this work, we pursue aligning training settings between binary and full-precision (FP) transformers in both pretraining and finetuning phases, which is helpful to bridge the training gap between binary and FP transformers.

We keep the pretraining settings of BiPFTs similar to BERTs. In detail, we train the same architectured binary BERT models in the base size with 110M parameters. We quantize weights and embeddings in transformers to the bit-width of 1 and quantize activations to the bit-width of 1 and 2 respectively. In pretraining, we use the BooksCorpus \cite{zhu2015aligning} and English Wikipedia \cite{devlin2018bert} as training data, including 800M and 2500M words respectively. The same as BERTs, lists, tables, and headers are ignored in Wikipedia.
In preprocessing, we follow the BERT and use the WordPiece tokenizer \cite{devlin2018bert} with a 30522 vocabulary size. The max length of each sentence is set to 128 tokens. And the batch size is set to 512 in one step. There are total $5\times10^5$ steps in pretraining which include about 3 epochs of all data. The same as full-precision conditions, we train binary models with an AdamW optimizer with a $2\times10^{-4}$ peak learning rate and 0.01 weight decay. A linear learning rate schedular with 5000 steps warm-up is also used. Our experiments show these common hyperparameters for most full-precision pretraining BERTs are general and robust enough for binary transformer pretraining.

In downstream tasks, we use the GLUE benchmark \cite{wang2018glue} to evaluate NLU performance. There are 8 subsets including CoLA, STS-B, MRPC, RTE, QQP, MNLI, QNLI. 
In finetuning, we also keep the same FP settings. In detail, we keep a constant $2\times10^{-5}$ learning rate and 32 batchsize for all the subsets, and we keep the same training epochs and evaluation settings as BiBERTs and BiTs. Notice that, we don't adapt to the best learning rate or batchsize for GLUE subsets like previous state-of-the-art works, which can improve performance a lot for BNNs but may result in overestimation of performance given new tasks.

\begin{figure}[t]
\centering
\includegraphics[width=1\columnwidth]{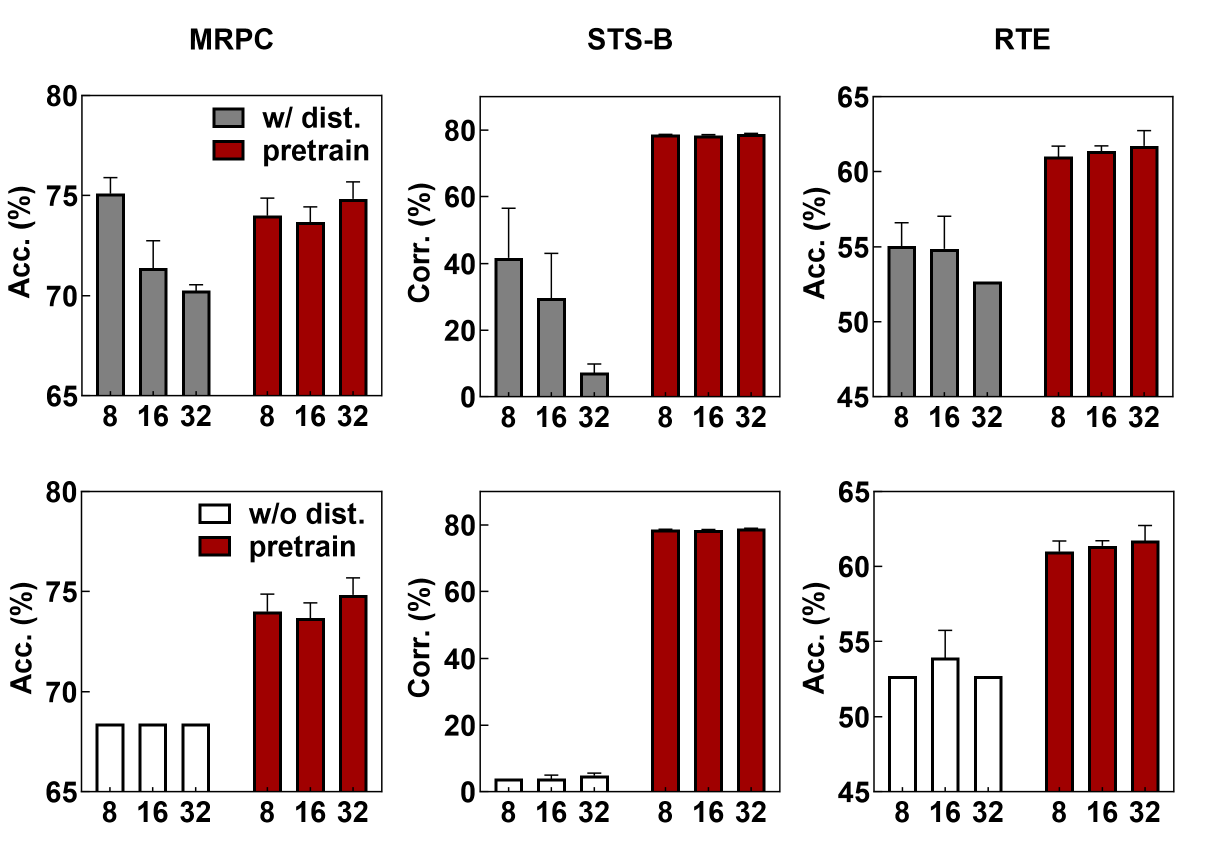}
\caption{Comparisons of BiPFT-A and baselines in different batch sizes. Up: baseline with task-specific distillation; down: baseline without task-specific distillation.}
\label{fig1}
\end{figure}

\begin{figure}[t]
\setlength{\abovecaptionskip}{0.cm}
\centering
\includegraphics[width=1\columnwidth]{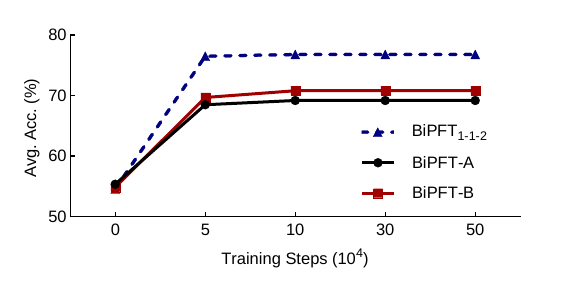}
\caption{Pertraining performance in different training steps.}
\label{fig5}
\end{figure}

\subsection{Main Results}

\begin{figure}[t]
\centering
\includegraphics[width=1\columnwidth]{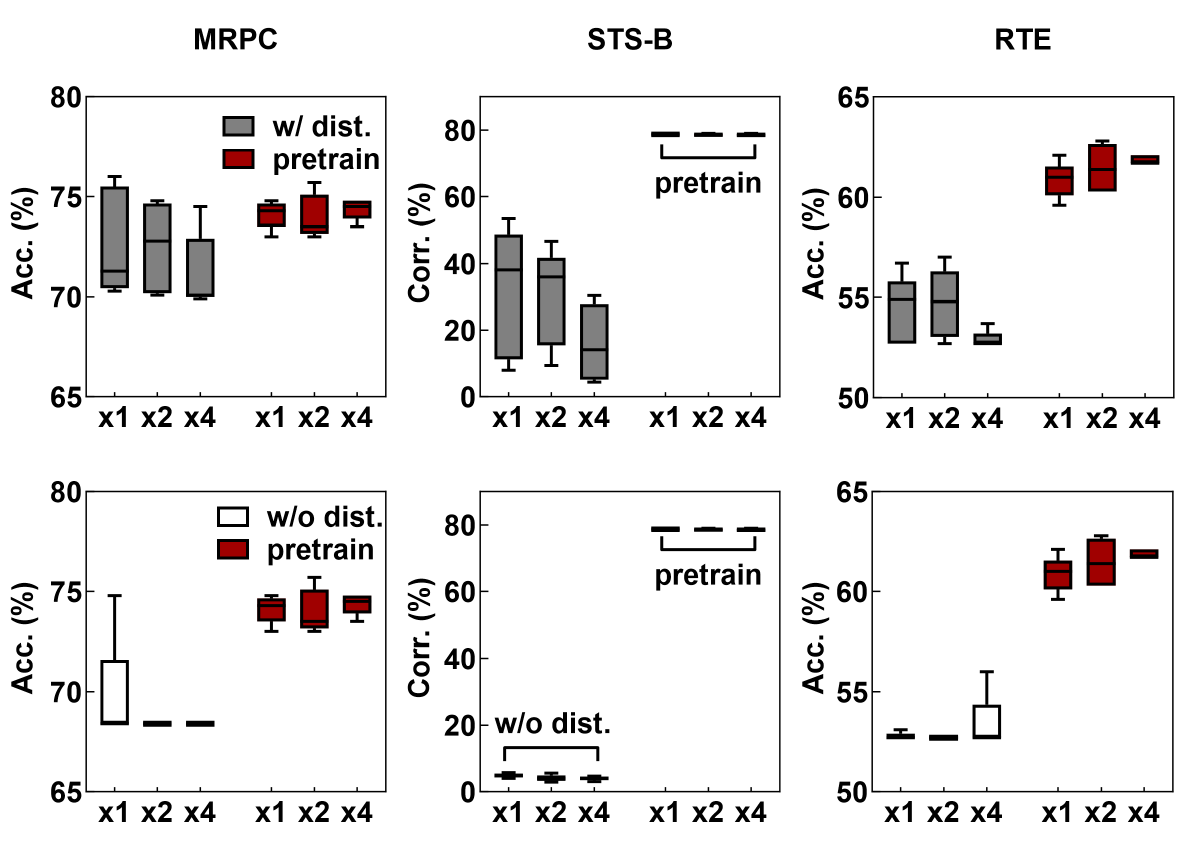}
\caption{Comparisons of BiPFT-A and baselines in different learning rates. Up: baseline with task-specific distillation; down: baseline without task-specific distillation. We set the base learning rates for baselines according to searched results of BiTs for every task; we set learning rates for BiPFT-A from $\{5\times10^\text{-6}, 1\times10^\text{-5}, 2\times10^\text{-5}\}$.}
\label{fig2}
\end{figure}

Table \ref{glue} shows comparisons with previous state-of-the-art BERTs in some low-bit quantization and binary. More detailed robustness and pretraining analysis are reported in Fig. \ref{fig1}, \ref{fig5}, \ref{fig2} respectively.

\noindent\textbf{Performance of BiPFT-A.} 
We summarize methods in Table \ref{methods}, where the baseline, baseline$^*$ and BiPFT-A share the same architecture; additional estimators are attached in BiPFT-B. 

We use baseline$^{*}$ to implement the previous task-specific pipeline in \textbf{Fig. \ref{figure1} (a)}. Baseline$^*$ uses the same hyperparameters searched by BiTs and is trained with a distillation, while evaluated in the common settings. Baseline$^{*}$ is competitive to surpass BiBERT by 4.9\% on average. 

To evaluate binary BERT performance in general settings, we remove task-specific hyperparameter search and distillation. The baseline model withdraws 12.8\% average accuracy dramatically, which indicates the weakness of binary BERT itself. This indicates the performance heavily relies on special training settings in task-specific binary BERTs.

After pretraining, BiPFT-A improves 13.9\% compared with baseline; even if compared with baseline$^{*}$ with additional distillation and hyperparameter search, BiPFT-A surpasses 1.1\%. This is the first time BNNs get rid of FP teachers and achieve better accuracy, which indicates pretraining significantly improves the learning ability of BNNs.

\noindent\textbf{Performance of BiPFT-B.} 
We report the performance of BiPFT-B in Table \ref{glue}. With the estimation of binarization residual polynomials, BiPFT-B further improves 1.6\% average performance compared with BiPFT-A. This indicates a large amount of pretraining data helps BNNs learn how to binarization in downstream. In total, the binary pretrained foundation model exceeds 15.4\% average performance compared with baseline, which narrows 57.2\% performance gap from the binary baseline to the FP BERT. In the setting of 2-bit activations, we also observe higger performance.

\begin{table*}[t]
\begin{center}
\renewcommand\arraystretch{0.7}
\centering
\setlength{\tabcolsep}{1.4mm}
\fontsize{10pt}{\baselineskip}\selectfont
\begin{tabular}{lcccccccccccc}
\hline
\textbf{Model} & \begin{tabular}[c]{@{}c@{}}\textbf{KQ$\uparrow$}\end{tabular} & \begin{tabular}[c]{@{}c@{}}\textbf{AttV$\uparrow$}\end{tabular} & \begin{tabular}[c]{@{}c@{}}\textbf{LoRA}\end{tabular} & \textbf{QQP} & \textbf{QNLI} & \textbf{SST-2} & \textbf{CoLA} & \textbf{STS-B} & \textbf{MRPC} & \textbf{RTE} & \begin{tabular}[c]{@{}c@{}}\textbf{MNLI}$_\text{-m/mm}$\end{tabular} & \textbf{Avg.} \\ 
\hline
FP-BERT$_{base}$     & -- & -- & --                  & 91.4 & 91.3 & 92.9 & 55.8 & 88.8 & 86.5 & 66.8 & 83.4/83.6 & 82.3 \\
\hdashline[0.8pt/1pt]
\multicolumn{5}{l}{\textit{W/o pretraining}} &&&&&&&& \\
{Baseline} & \XSolidBrush & \XSolidBrush       & --  & 78.0 & 60.2 & 79.7 & 16.5 & 19.7 & 70.3 & 57.0 & 58.8/58.7 & 55.4 \\
{Baseline} & \Checkmark & \Checkmark           & --  & 78.2 & 60.6 & 79.5 & 11.1 & 19.5 & 70.3 & 55.6 & 59.5/59.7 & 54.9 \\
\hdashline[0.8pt/1pt]
\multicolumn{5}{l}{\textit{Ablation in architectures}} &&&&&&&& \\
{BiPFT-A} & \XSolidBrush & \XSolidBrush & --         & 82.8 & 80.3 & 85.6 & 19.8 & 79.0 & 74.0 & 60.6 & 70.3/70.8 & 69.2 \\
{BiPFT-A} & \XSolidBrush & \Checkmark & --           & 84.0 & 81.5 & 85.9 & 19.4 & 79.4 & 73.8 & 63.2 & 70.5/71.0 & 69.9 \\
{BiPFT-A} & \Checkmark & \XSolidBrush & --           & 83.7 & 81.7 & 85.7 & 21.3 & 79.3 & 76.2 & 62.8 & 70.0/70.9 & 70.2 \\
{BiPFT-B} & \Checkmark & \Checkmark & --             & 83.7 & 81.7 & 86.2 & 22.9 & 80.2 & 76.2 & 66.1 & 69.5/70.6 & 70.8 \\
\hdashline[0.8pt/1pt]
\multicolumn{5}{l}{\textit{Ablation in ranks}} &&&&&&&& \\
{BiPFT-B-rank1} & \Checkmark & \Checkmark & --       & 83.7 & 81.7 & 86.2 & 22.9 & 80.2 & 76.2 & 66.1 & 69.5/70.6 & 70.8 \\
{BiPFT-B-rank2} & \Checkmark & \Checkmark & --       & 83.3 & 80.9 & 87.3 & 18.7 & 78.8 & 73.5 & 65.3 & 69.4/70.4 & 69.7 \\
{BiPFT-B-rank4} & \Checkmark & \Checkmark & --       & 83.4 & 81.6 & 86.2 & 18.9 & 76.9 & 75.0 & 61.0 & 69.7/70.4 & 69.2 \\
\hdashline[0.8pt/1pt]
\multicolumn{5}{l}{\textit{Comparison with LoRA}} &&&&&&&& \\
{BiPFT-A} & \XSolidBrush & \XSolidBrush & \Checkmark & 83.5 & 81.4 & 85.3 & 18.1 & 80.7 & 76.5 & 62.8 & 70.1/70.9 & 69.9 \\
{BiPFT-B} & \Checkmark & \Checkmark & \Checkmark     & 81.9 & 79.3 & 83.9 & 16.5 & 79.4 & 73.3 & 59.9 & 68.3/69.8 & 68.0 \\
{BiPFT-B} & \Checkmark & \Checkmark & \XSolidBrush   & 83.7 & 81.7 & 86.2 & 22.9 & 80.2 & 76.2 & 66.1 & 69.5/70.6 & 70.8 \\
\hline
\end{tabular}
\end{center}
\caption{Ablation studies for BiPFTs. KQ$\uparrow$ indicates adding estimators for key and query as Eq. \ref{eq_18}; AttV$\uparrow$ indicates adding estimators for value as Eq. \ref{eq_27}.}
\label{ablation}
\end{table*}

In RTE, MRPC and STS-B datasets, there are 2.5k, 3.7k and 7k data respectively and they are relatively small datasets in GLUE. We observe BiPFT-B has more significant improvements than relatively big subsets, which are 9.1\%, 5.9\% and 60.5\% in RTE, MRPC and STS-B. Even if compared with distilled baseline$^{*}$, BiPFT-B surpasses 9.4\%, 0.2\% and 26.8\% on average respectively. This indicates knowledge distillation is hard to make up the performance drop caused by the missing binary pretrained foundation models, in small downstream datasets.

\noindent\textbf{Efficiency analysis.} In Table \ref{glue}, we compare operations and memory usage between FP and low-bit models. Compared with FP BERTs in base size, BiPFTs-B saves 56$\times$ operations and 28$\times$ memory for the 1-bit activations, while saves 28$\times$ operations and 28$\times$ memory for 2-bit activations.

\noindent\textbf{Robustness analysis of binary transformers.} 
We select three datasets, RTE, MRPC, STS-B and analyze robustness in different training settings. 
In Fig. \ref{fig1} and \ref{fig2}, we evaluate STS-B by the average of Pearson and Spearman correlation.
In Fig. \ref{fig1}, we compare BiPFT-A and baseline models in different batchsize settings to evaluate the robustness of pretraining in different batchsizes. 
In Fig. \ref{fig2}, we compare BiPFT-A and baselines in different learning rate settings to evaluate the robustness of pretraining in different learning rates.
More detailed results are shown in Appendix B of our extended version \cite{xing2023bipft}.

Our observations are mainly in three aspects. 
Firstly, in Fig. \ref{fig1} (up), compared with baseline$^*$ with distillation, BiPFT-A keeps almost higher performance stably in different batchsizes. Although task-specific distillation achieves at most 76.0\% acc. in MRPC in batchsize 8, performance drops dramatically when training batchsize increases. The small batchsize makes it more challenging for parallel computation.
Secondly, in Fig. \ref{fig2} (up), we train baseline$^*$ and BiPFT-A in different learning rates and evaluate the variance of results. In different learning rates, pretraining helps to stabilize performance significantly. Because of unstable performance, previous binary transformers have to perform hyperparameter search for different tasks, which can be inefficient and unstable.
Thirdly, binary transformers heavily rely on distillation. When without pretraining, there are weak learning capabilities as shown in Fig \ref{fig1} (down), \ref{fig2} (down). In MRPC dataset, binary classification accuracy directly drops to 68.4\% which is similar to encounter model degeneration or random choice; in STS-B dataset, the Pearson and Spearman correlation close to 0\%. These phenomena indicate task-specific binary transformers have a high risk to lost learning ability once removing FP teachers. 

\noindent\textbf{Pretraining time analysis.} Fig. \ref{fig5} shows the average GLUE performance of BiPFTs in different pretraining steps. In early pretraining time, downstream performance improves with more training steps. For the base-sized binary BERTs with 110M binary parameters, $1\times10^{5}$ pretraining steps are enough for fully pretraining, where the batch size is 512. 
This confirms enough training time for binary transformers.

\subsection{Ablation Studies}
Table \ref{ablation} shows ablation studies for BiPFTs.
More ablations for initialization are shown in Appendix C of our extended version \cite{xing2023bipft}.

\noindent\textbf{Ablation in architectures.} 
In BiPFT-B, we estimate binarization residual polynomials in two steps according to Eq. \ref{eq_18}, \ref{eq_27} respectively. 
As shown in Table \ref{ablation}, with pretraining, it improves average performance when using estimators in Eq. \ref{eq_18} or Eq. \ref{eq_27} alone.
When combining the estimations in both Eq. \ref{eq_18} and \ref{eq_27} together, it carries out the best and totally improves 1.6\% performance on average. 
This confirms that low-rank multiplications have the capacity to learn to estimate binarization residual polynomials from queries, keys and values accordingly. 
However,data-driven binarization polynomial estimators are data-hungry. When without pretraining, estimators can't achieve better performance.

\noindent\textbf{Ablation in ranks.} 
We use the low-rank matrix multiplications as binarization polynomial estimators. By default, we use rank number 1 to reduce computational cost. To investigate the influence of ranks, We revise the rank number in Eq. \ref{eq_18} as 2 and 4. In Table \ref{ablation}, increasing the rank can't improve performance, which indicates larger ranks may encounter overfitting to binarization residual polynomials.

\noindent\textbf{Comparison with LoRA.} 
Because we use low-rank binarization estimators to improve binary multiplications between queries, keys and values, one potential idea could be whether we can use LoRA \cite{hu2022lora} to improve linear layers in self-attention. As shown in Table \ref{ablation}, although adding LoRA to binary transformers alone improves results, LoRA is less efficient compared with low-rank estimators of binarization residual polynomials. Moreover, when we both add LoRA and binarization polynomial estimators, it encounters unstable performance in our experiments, because of overfitting of low-rank parameters. As a result, low-rank estimators of binarization polynomials are more efficient and explicable for binary self-attention. 

\section{Conclusion}

This work proposes the first binary pretrained foundation model for NLU tasks, promoting BNNs to the era of pretraining.
This provides a lot of conveniences to finetune accurate, robust and training efficient binary transformers in downstream tasks.
In the future, we think it would be meaningful to pretrain binary foundation models for natural language generation (NLG) tasks like current GPT \cite{brown2020language} and LLama \cite{touvron2023llama}, instead of downstream binary models. General knowledge is able to significantly improve the learning capabilities of BNNs.

\bibentry{c:22}

\section{Acknowledgments}
This work is supported by the National Key R\&D Program of China (No.2022ZD0116301) and the National Science Foundation of China under grant No.62206150. This work is also supported by NSFC No.62106249.

\bibliography{aaai24}
\clearpage

\section{Appendix}
\subsection{A. Discussion for Baseline Settings}

We define the baseline binary transformer as a benchmark to explore the general performance of binarization. Our key principles include:
\begin{itemize}
\item generally used binarization settings and model architectures (for example, activation functions);
\item common and simple training strategies;
\item common and fair evaluation settings;
\item state-of-the-art performance.
\end{itemize}

We make detailed comparisons between BiTs and our baseline binary transformers as shown in Table \ref{table1}. Our necessary modifications from the original BiTs include:
\begin{itemize}
\item we use the same binarization level \{-1,+1\} in FFNs. BiTs use different binarization levels between activations and weights in FFNs. In our baselines, we utilize common binarization settings as shown in Table \ref{table1};
\item we start with defining the baseline{$^*$} training setting the same as BiTs, which is equipped with grid-searched hyperparameters and task-specific distillation. Furthermore, we remove grid-searched hyperparameters and task-specific distillation and define the baseline training setting, which is the same as the finetuning of BiPFTs;
\item we use the common evaluation settings as full-precise BERTs. Previous binary BERTs often use a frequent evaluation and report the best validation performance because of the unstability in training. This improves the risk of overfitting validation sets. 
\end{itemize}

\subsection{B. Robustness of Binary Transformers}

We report detailed results of the robustness of binary transformers in Table. \ref{search}, \ref{search2}, and \ref{search3}.

In Table \ref{search}, we explore the robustness of the baseline$^*$ in conditions of both with and without task-specific distillation. In detail, we train baseline$^*$ in different learning rates and batchsizes. The base learning rate in baseline$^*$ is set according to the searched results of BiTs. A linear strategy between batchsizes and learning rates is also applied to determine the range of learning rates. For example, if the base learning rate is $5\times10^\texttt{-4}$ for batchsize 8, the max learning for batchsize 32 is set to $2\times10^\texttt{-3}$. We finally report the statistical results of Table \ref{search} in Fig. 2 and 4 in this paper.

In Table \ref{search2}, we also report the robustness of the baseline$^*$ in the learning rate of $2\times10^\texttt{-5}$, which is used as the unified learning rate of BiPFTs and BERTs in finetuning. In general, performance of baseline$^*$ drops in this more general setting compared with the searched learning rate in Table \ref{search}.

\begin{table}
\centering
\resizebox{\linewidth}{!}{
\begin{tabular}{l|cc|ccccc}
\hline
data & DT & LR & bs-8 & bs-12 & bs-16 & bs-24 & bs-32 \\
\hline
MRPC & \XSolidBrush & $5\times10^\texttt{-4}$ & 74.8&68.4&68.4&68.4&68.4  \\
MRPC & \XSolidBrush & $1\times10^\texttt{-3}$ & 68.4&68.4&68.4&68.4&68.4  \\
MRPC & \XSolidBrush & $2\times10^\texttt{-3}$ & 68.4&68.4&68.4&68.4&68.4  \\
\hdashline[0.8pt/1pt]
STS-B & \XSolidBrush & $5\times10^\texttt{-4}$ & 4.0 & 5.7 & 5.1 & 4.6 & 4.8  \\
STS-B & \XSolidBrush & $1\times10^\texttt{-3}$ & 4.0 & 2.9 & 3.6 & 4.2 & 5.6  \\
STS-B & \XSolidBrush & $2\times10^\texttt{-3}$ & 4.0 & 4.0 & 3.0 & 4.7 & 4.3  \\
\hdashline[0.8pt/1pt]
RTE & \XSolidBrush & $5\times10^\texttt{-4}$ & 52.7&52.7&53.1&52.7&52.7  \\
RTE & \XSolidBrush & $1\times10^\texttt{-3}$ & 52.7&52.7&52.7&52.7&52.7  \\
RTE & \XSolidBrush & $2\times10^\texttt{-3}$ & 52.7&52.7&56.0&52.7&52.7  \\
\hline
MRPC & \Checkmark & $5\times10^\texttt{-4}$ & 76.0&75.0&71.3&70.3&70.6 \\
MRPC & \Checkmark & $1\times10^\texttt{-3}$ & 74.8&74.5&72.8&70.3&70.1 \\
MRPC & \Checkmark & $2\times10^\texttt{-3}$ & 74.5&71.3&70.1&69.9&70.1 \\
\hdashline[0.8pt/1pt]
STS-B & \Checkmark & $5\times10^\texttt{-4}$ & 53.4 & 44.0 & 38.1 & 14.6 & 8.0 \\
STS-B & \Checkmark & $1\times10^\texttt{-3}$ & 46.7 & 36.0 & 36.7 & 21.5 & 9.4 \\
STS-B & \Checkmark & $2\times10^\texttt{-3}$ & 25.1 & 30.5 & 14.1 & 5.9  & 4.4 \\
\hdashline[0.8pt/1pt]
RTE & \Checkmark & $5\times10^\texttt{-4}$ & 56.7&54.9&54.9&52.7&52.7 \\
RTE & \Checkmark & $1\times10^\texttt{-3}$ & 54.8&55.6&57.0&53.4&52.7 \\
RTE & \Checkmark & $2\times10^\texttt{-3}$ & 53.7&52.7&52.7&52.7&52.7 \\
\hline
\end{tabular}
}
\caption{Performance of baseline$^*$ models in different training settings, where DT, LR, and bs indicate the task-specific distillation, learning rate and batchsize respectively.}
\label{search}
\end{table}

\begin{table}
\centering
\resizebox{\linewidth}{!}{
\begin{tabular}{l|cc|ccccc}
\hline
data & DT & LR & bs-8 & bs-12 & bs-16 & bs-24 & bs-32 \\
\hline
MRPC & \XSolidBrush & $2\times10^\texttt{-5}$ & 68.4&68.9&68.6&68.9&68.9  \\
MRPC & \Checkmark & $2\times10^\texttt{-5}$ & 70.3&68.9&68.6&68.4&68.4  \\
\hdashline[0.8pt/1pt]
STS-B & \XSolidBrush & $2\times10^\texttt{-5}$ & 18.4&16.7&12.5&9.6&9.8  \\
STS-B & \Checkmark & $2\times10^\texttt{-5}$ & 9.3&4.9&5.0&5.9&5.0  \\
\hdashline[0.8pt/1pt]
RTE & \XSolidBrush & $2\times10^\texttt{-5}$ & 55.6&57.8&54.9&56.0&54.9 \\
RTE & \Checkmark & $2\times10^\texttt{-5}$ & 58.8&54.9&55.6&53.1&54.5  \\
\hline
\end{tabular}
}
\caption{Performance of baseline$^*$ models in different training settings, where DT, LR, and bs indicate the task-specific distillation, learning rate and batchsize respectively. In constrast to using a searched base learning rate from BiTs (Table \ref{search}), this table reports results of using the same learning rate as BiPFT-A.}
\label{search2}
\end{table}

\begin{table}
\centering
\resizebox{\linewidth}{!}{
\begin{tabular}{l|cc|ccccc}
\hline
data & DT & LR & bs-8 & bs-12 & bs-16 & bs-24 & bs-32 \\
\hline
MRPC & \XSolidBrush & $2\times10^\texttt{-5}$ & 74.5&74.3&73.0&74.8&74.0  \\
MRPC & \XSolidBrush & $1\times10^\texttt{-5}$ & 73.0&73.5&74.5&73.3&75.7  \\
MRPC & \XSolidBrush & $5\times10^\texttt{-6}$ & 74.5&74.3&73.5&74.8&74.8  \\
\hdashline[0.8pt/1pt]
STS-B & \XSolidBrush & $2\times10^\texttt{-5}$ & 78.7&78.7&78.7&78.6&78.9  \\
STS-B & \XSolidBrush & $1\times10^\texttt{-5}$ & 78.5&79.0&78.3&78.3&78.7  \\
STS-B & \XSolidBrush & $5\times10^\texttt{-6}$ & 78.6&78.4&78.3&78.4&79.0  \\
\hdashline[0.8pt/1pt]
RTE & \XSolidBrush & $2\times10^\texttt{-5}$ & 61.0&62.1&61.0&59.6&60.6  \\
RTE & \XSolidBrush & $1\times10^\texttt{-5}$ & 60.3&60.3&61.4&62.8&62.5  \\
RTE & \XSolidBrush & $5\times10^\texttt{-6}$ & 61.7&61.7&61.7&62.1&62.1  \\
\hline
\end{tabular}
}
\caption{Performance of BiPFT-A models in different training settings, where DT, LR, and bs indicate the task-specific distillation, learning rate and batchsize respectively.}
\label{search3}
\end{table}

\begin{table*}[!htbp]
\caption{Definition of our baseline models, where 'Down. Dist.' indicates downstream distillation; $ \{-1, +1\} $ and $ \{0, +1\} $ indicate different binarization levels. We compare the binarization, training, and evaluation difference of binary transformers \cite{bai-etal-2021-binarybert, qin2022bibert, liu2022bit}. We keep the general settings (in \textcolor{red}{red}) the same as the full-precision (FP) BERT or the most common choices.}
\label{table1}
\begin{center}
\resizebox{\linewidth}{!}{
\begin{small}
\begin{tabular}{l|ccccc|cc|r}
\hline
\textbf{Method}   & \textbf{Activations} & \textbf{Weights} & \textbf{Self-Atten.} & \textbf{FFN Act.} & \textbf{Act. Func.} & \textbf{Lr/Batchsize} & \textbf{Down. Dist.} & \textbf{Eval Freq.} \\
\hline
FP-BERT &  -- & -- & -- & -- & GeLU & $2\times10^\texttt{-5}/32 $  & \XSolidBrush & 1 Epoch     \\
BinaryBERT     & $ \{-1, +1\} $ & $ \{-1, +1\} $ & $ \{-1, +1\} $ & {$ \{-1, +1\} $} 				& {GeLU} 				 & \textcolor{blue}{Tuning} & \textcolor{blue}{\Checkmark} & \textcolor{blue}{100 Steps} \\
BiBERT         & $ \{-1, +1\} $ & $ \{-1, +1\} $ & $ \{-1, +1\} $ & {$ \{-1, +1\} $} 				& {GeLU} 				 & \textcolor{blue}{Tuning} & \textcolor{blue}{\Checkmark} & \textcolor{blue}{200 Steps} \\
BiT            & $ \{-1, +1\} $ & $ \{-1, +1\} $ & $ \{-1, +1\} $ & \textcolor{blue}{$ \{0, +1\} $} & \textcolor{blue}{ReLU} & \textcolor{blue}{Grid Search} & \textcolor{blue}{\Checkmark} & \textcolor{blue}{100 Steps} \\
baseline$^{*}$ & $ \{-1, +1\} $	& $ \{-1, +1\} $ & $ \{-1, +1\} $ & \textcolor{red}{$ \{-1, +1\} $} & \textcolor{red}{GeLU}  & \textcolor{blue}{Grid Search} & \textcolor{blue}{\Checkmark} & \textcolor{red}{1 Epoch} \\
baseline       & $ \{-1, +1\} $	& $ \{-1, +1\} $ & $ \{-1, +1\} $ & \textcolor{red}{$ \{-1, +1\} $} & \textcolor{red}{GeLU}  & \textcolor{red}{$2\times10^\texttt{-5}/32 $} & \textcolor{red}{\XSolidBrush} & \textcolor{red}{1 Epoch} \\
\hline
\end{tabular}
\end{small}
}
\end{center}
\end{table*}

\begin{table*}[!htbp]
\begin{center}
\renewcommand\arraystretch{0.6}
\centering
\caption{Ablation studies for initialization. FP-init indicates initializing from a full-precision task-specific BERT for baseline$^*$; it indicates initializing from full-precision pretrained BERT for BiPFT-B. Different with Table \ref{glue}, \ref{ablation}, SST-B results are evaluated by the average of Pearson and Spearman correlation.}
\label{ablation2}
\setlength{\tabcolsep}{1.4mm}
\fontsize{10pt}{\baselineskip}\selectfont
\begin{tabular}{lccccccccccc}
\hline
\textbf{Model} & \textbf{FP-init.} & \textbf{QQP} & \textbf{QNLI} & \textbf{SST-2} & \textbf{CoLA} & \textbf{STS-B} & \textbf{MRPC} & \textbf{RTE} & \begin{tabular}[c]{@{}c@{}}\textbf{MNLI}$_\text{-m/mm}$\end{tabular} & \textbf{Avg.} \\ 
\hline
FP-BERT     & --        & 91.4 & 91.3 & 92.9 & 55.8 & 89.0 & 86.5 & 66.8 & 83.4/83.6 & 82.3 \\
\hdashline[0.8pt/1pt]
{baseline$^*$} & \XSolidBrush    & 85.1 & 83.3 & 83.8 & 21.2 & 5.3  & 70.6 & 56.7 & 73.6/74.4 & 61.6 \\
{baseline$^*$} & \Checkmark      & 85.5 & 84.2 & 85.7 & 21.9 & 53.4 & 76.0 & 56.7 & 74.8/75.4 & 68.1 \\
\hdashline[0.8pt/1pt]
{BiPFT-B} & \XSolidBrush         & 84.6 & 81.3 & 84.9 & 18.7 & 79.5 & 75.0 & 60.6 & 70.7/71.3 & 69.6 \\
{BiPFT-B} & \Checkmark           & 83.7 & 81.7 & 86.2 & 22.9 & 80.4 & 76.2 & 66.1 & 69.5/70.6 & 70.8 \\
\hline
\end{tabular}
\end{center}
\end{table*}

In Table \ref{search3}, as a comparation of Table \ref{search}, we also report the robustness of BiPFT-A. BiPFT-A and baseline$^*$ have the same architecture but use different training strategies. We use common learning rate $2\times10^\texttt{-5}$ for batchsize 32. According the linear stragety between learning rates and batchsizes, the minimum learning rate for batchsize 8 is set to $5\times10^\texttt{-6}$. Finally, we also report the statistical results in Fig. 2 and 4 in this paper. It is obvious that the robustness of BiPFT-A largely improved compared with baseline$^*$ in different learning rates and batchsizes.

\subsection{C. Ablations in Initialization}

We make comparations about whether to use a random initialization or initializing from full-precision models. In baseline$^*$, it is initialized from a finetuned full-precision BERT in every task. In BiPFT-B, it is initialized from a pretrained foundation BERT in the HuggingFace. As shown in Table \ref{ablation2}, initializing from full-precision models helps improve final performance in both baseline$^*$ and BiPFT-B. In contrast to downstream models, initializing from a foundation model in the pretraining stage doesn't influence the finetuning given new tasks.

\end{document}